%% file: arxiv.tex
\documentclass[journal]{IEEEtran}
\usepackage{amsmath,amsfonts}
\usepackage{url}
\usepackage{marvosym}
\usepackage{balance}
\usepackage[hidelinks,hypertexnames=false]{hyperref}
\usepackage{graphicx}
\usepackage{cite}
\usepackage{booktabs}
\usepackage{multirow}
\usepackage{threeparttable}
\usepackage{makecell}
\usepackage{mathrsfs}
\usepackage{enumitem}
\usepackage{ulem}
\usepackage{color}
\usepackage{float}
\usepackage{caption}
\input{packages.tex}
\newcolumntype{L}[1]{>{\raggedright\arraybackslash}p{#1}}
\newcolumntype{C}[1]{>{\centering\arraybackslash}p{#1}}
\newcolumntype{R}[1]{>{\raggedleft\arraybackslash}p{#1}}
\DeclareMathOperator*{\maxc}{max}
\definecolor{TYS}{rgb}{0.6, 0.8, 0.2}
\definecolor{DOcolor}{rgb}{1,0.45,0.0}
\definecolor{NAVYcolor}{rgb}{0.05,0,0.5}

\begin{document}
	\title{S$^3$M-Net: Joint Learning of Semantic Segmentation and Stereo Matching for Autonomous Driving}
	
	\normalem
	\author{Zhiyuan Wu$^{\orcidicon{0009-0001-7253-7603}\,}$, Yi Feng$^{\orcidicon{0009-0005-4885-0850}\,}$, Chuang-Wei Liu$^{\orcidicon{0000-0003-0260-6236}\,}$, Fisher Yu$^{\orcidicon{0000-0001-8829-7344}\,}$, \\ Qijun Chen$^{\orcidicon{0000-0001-5644-1188}\,}$,~\IEEEmembership{Senior Member,~IEEE}, and Rui Fan$^{\orcidicon{0000-0003-2593-6596}\,}$,~\IEEEmembership{Senior Member,~IEEE}
		\vspace{-2em}
		\thanks{
			This research was supported by the Science and Technology Commission of Shanghai Municipal under Grant 22511104500, the National Natural Science Foundation of China under Grant 62233013, the Fundamental Research Funds for the Central Universities, and Xiaomi Young Talents Program. (\emph{Corresponding author: Rui Fan})}
		\thanks{Zhiyuan Wu, Yi Feng, Chuang-Wei Liu, Qijun Chen, and Rui Fan are with the College of Electronics \& Information Engineering, Shanghai Research Institute for Intelligent Autonomous Systems, the State Key Laboratory of Intelligent Autonomous Systems, and Frontiers Science Center for Intelligent Autonomous Systems, Tongji University, Shanghai 201804, China (e-mails: gwu@tongji.edu.cn, fengyi@ieee.org, \{cwliu, qjchen, rfan\}@tongji.edu.cn).}
		\thanks{Fisher Yu is with the Department of Information Technology and Electrical Engineering, ETH Zürich, Sternwartstrasse 7, 8092 Zürich, Switzerland. email: i@yf.io.}
	}
	
	\markboth{IEEE Transactions on Intelligent Vehicles}{}
	
	\maketitle
	
	\begin{abstract}
		Semantic segmentation and stereo matching are two essential components of 3D environmental perception systems for autonomous driving. Nevertheless, conventional approaches often address these two problems independently, employing separate models for each task. This approach poses practical limitations in real-world scenarios, particularly when computational resources are scarce or real-time performance is imperative. Hence, in this article, we introduce S$^3$M-Net, a novel joint learning framework developed to perform semantic segmentation and stereo matching simultaneously. Specifically, S$^3$M-Net shares the features extracted from RGB images between both tasks, resulting in an improved overall scene understanding capability. This feature sharing process is realized using a feature fusion adaption (FFA) module, which effectively transforms the shared features into semantic space and subsequently fuses them with the encoded disparity features. The entire joint learning framework is trained by minimizing a novel semantic consistency-guided (SCG) loss, which places emphasis on the structural consistency in both tasks. Extensive experimental results conducted on the vKITTI2 and KITTI datasets demonstrate the effectiveness of our proposed joint learning framework and its superior performance compared to other state-of-the-art single-task networks. Our project webpage is accessible at mias.group/S3M-Net.
	\end{abstract}
	
	\begin{IEEEkeywords}
		semantic segmentation, stereo matching, environmental perception, autonomous driving, joint learning.
	\end{IEEEkeywords}
	
	\section{Introduction}
	\label{sec.intro}
	
	\IEEEPARstart{3}{D} environmental perception stands as a critical and foundational aspect of autonomous driving \cite{ranft2016tiv, fisher2016tiv}. Semantic segmentation and stereo matching are two key functionalities in 3D environmental perception systems \cite{fan2020sneroadseg, xu2023igevstereo, rao2023masked}. The former provides a comprehensive pixel-level understanding of the environment, while the latter simulates human binocular vision to acquire accurate and dense depth information \cite{fan2023autonomous}. The combined utilization of both functionalities has become the mainstream approach in recent years \cite{yang2018segstereo, zhang2019dispsegnet, wu2019semantic, zhan2019dsnet, dovesi2020real, chen2020sgnet}.
	
	In recent years, the research focus in semantic segmentation has shifted from single-modal networks \cite{badrinarayanan2017segnet, ronneberger2015unet, zhao2017pspnet, chen2017deeplabv3, chen2018deeplabv3plus, huang2019ccnet} with a single encoder to feature-fusion networks with dual encoders \cite{hazirbas2017FuseNet, sun2019RTFNet, wang2021sne, min2022orfd}. The latter type of networks extract heterogeneous features from RGB-X data, where ``X'' can represent various forms of spatial geometric information, \textit{e.g.}, depth images generated from LiDAR point clouds and surface normal maps obtained through depth-to-normal translation \cite{yang2023three}. These heterogeneous features are subsequently fused to achieve a more comprehensive understanding of the environment \cite{wang2021sne}. However, a critical drawback of feature-fusion networks is their dependency on the availability of the ``X'' data, which can pose limitations in scenarios where LiDARs are not present. Additionally, when the accuracy of the ``X'' data is not satisfactory, such as due to variations in camera-LiDAR calibration, the fusion of these heterogeneous features can potentially lead to a degradation in the overall performance of semantic segmentation \cite{li2023roadformer}. While a stereo camera can serve as a practical and cost-effective alternative to LiDARs for depth information acquisition, the incorporation of a separate stereo matching network introduces additional computations, and therefore, poses difficulties in achieving real-time processing speeds for the entire system \cite{wu2019semantic}. Moreover, stereo matching and semantic segmentation share the same input, and the representations from RGB images can be more informative when they are jointly learned by both tasks.
	
	The joint learning of multiple interconnected 3D environmental perception tasks introduces a form of regularization that has demonstrated superiority over uniform complexity penalization in reducing over-fitting \cite{fan2023one}. Furthermore, rather than employing separate models for semantic segmentation and stereo matching, joint learning can potentially reduce computational complexity \cite{yang2018segstereo, zhang2019dispsegnet, wu2019semantic, zhan2019dsnet, dovesi2020real, chen2020sgnet}, as shared learning representations can be used for both tasks. This can be advantageous in real-time or resource-constrained applications. Moreover, joint learning enables end-to-end optimization of the entire system, allowing the model to adapt to the specific challenges of both tasks simultaneously. Consequently, this can lead to improved performance when compared to models trained separately for each task \cite{chen2020sgnet}. In addition, stereo matching can occasionally produce ambiguously estimated disparities, particularly in texture-less or occluded regions \cite{fan2018road}. Semantic segmentation can provide informative contextual information that helps disambiguate such cases, ultimately leading to more reliable disparity estimations \cite{wu2019semantic}. Regrettably, the joint learning of semantic segmentation and stereo matching, especially within feature-fusion networks or when faced with a scarcity of training samples, has received relatively limited attention in this research area and calls for further investigation.
	
	Therefore, in this article, we present \uline{\textbf{S}emantic \textbf{S}egmentation and \textbf{S}tereo \textbf{M}atching \textbf{Net}work (\textbf{S$^3$M-Net})}, a joint framework to simultaneously predict both semantic and disparity information. S$^3$M-Net begins with the extraction of features from stereo images. These features are then processed by a multi-level gate recurrent unit (GRU) operator to generate a disparity map. Simultaneously, these features are shared with the semantic segmentation task via a feature fusion adaptation (FFA) module.  Building upon our prior work SNE-RoadSeg \cite{fan2020sneroadseg}, we extract additional features from the estimated disparity map. Finally, a densely-connected skip connection decoder is employed to decode the fused features and generate the semantic predictions. S$^3$M-Net is trained in a fully supervised manner by minimizing a semantic consistency-guided (SCG) joint learning loss. Extensive experiments conducted on the vKITTI2 \cite{cabon2020vkitti2} and KITTI 2015 \cite{menze2015kitti} datasets unequivocally demonstrate the effectiveness and superior performance of our proposed S$^3$M-Net.
	
	In summary, the main contributions of this article include: 
	\begin{itemize}
		\item S$^3$M-Net, a joint learning framework designed to address semantic segmentation and stereo matching simultaneously, where both tasks collaboratively leverage the features extracted from RGB images, enhancing the overall understanding of the driving scenario;
		\item A feature fusion adaption module to transform the shared feature maps into semantic space and subsequently fuse them with encoded disparity features;
		\item A semantic consistency-guided loss function to supervise the training process of the joint learning framework, emphasizing on the structural consistency in both tasks.
	\end{itemize}
	
	The remainder of this article is organized as follows: 
	Sect. \ref{sec.related_works} provides a review of related work.
	Sect. \ref{sec.methodology} introduces our proposed S$^3$M-Net.
	Sect. \ref{sec.experiments} presents the experimental results and compares our framework with other state-of-the-art (SoTA) approaches.
	In Sect. \ref{sec.discussion}, we discuss the advantages and limitations of our method. 
	Finally, we conclude this article in Sect. \ref{sec.conclusion}.

	\section{Literature Review}
	\label{sec.related_works}
	
	\subsection{Semantic Segmentation}
	\label{sec.semantic_segmentation}
	Semantic segmentation has been a long-standing problem in the field of computer vision over the past decade \cite{fan2023autonomous, michieli2020tiv}. The SoTA networks in this research area can generally be classified into two categories: (1) single-modal networks with a single encoder and (2) feature-fusion networks with multiple encoders \cite{fan2020sneroadseg, yang2023tiv, fan2022tiv}. In the early attempts to tackle semantic segmentation, researchers primarily focused on encoder-decoder architectures for pixel-level classification. Notable examples include SegNet \cite{badrinarayanan2017segnet}, U-Net \cite{ronneberger2015unet}, PSPNet \cite{zhao2017pspnet}, the DeepLab series \cite{chen2017deeplabv3, chen2018deeplabv3plus}, and Transformer-based networks \cite{strudel2021segmenter, xie2021segformer, cheng2022mask2former}. The encoder extracts hierarchical deep features from the input image, while the decoder produces the segmentation map by upsampling and combining the features from different encoder layers. However, these networks are limited in their ability to effectively combine deep features extracted from different modalities (or sources) of visual information. As a result, they often struggle to produce accurate segmentation results in challenging scenarios characterized by poor lighting and illumination conditions \cite{fan2020sneroadseg}. Therefore, researchers have turned their focus towards feature-fusion networks that can effectively integrate deep features learned from multiple modalities (or sources) of visual information. This problem is commonly referred to as ``RGB-X semantic segmentation'', where ``X'' represents the additional modality (or source) of visual information, in addition to the RGB images. The most representative feature-fusion networks based on convolutional neural networks (CNNs) include FuseNet \cite{hazirbas2017FuseNet}, MFNet \cite{ha2017MFNet}, RTFNet \cite{sun2019RTFNet}, and our previous works SNE-RoadSeg series \cite{fan2020sneroadseg, wang2021sne}. Furthermore, Transformer-based RGB-X semantic segmentation networks, such as OFF-Net \cite{min2022orfd} and RoadFormer \cite{li2023roadformer}, have been recently introduced. In this article, we design our S$^3$M-Net based on the SNE-RoadSeg architecture and explore more effective solutions for the feature fusion operation. 
	
	\subsection{Stereo Matching}
	\label{sec.stereo_matching}
	Conventional explicit programming-based stereo matching algorithms (local, global, and semi-global) generally consist of four main procedures: (1) cost computation, (2) cost aggregation, (3) disparity optimization, and (4) disparity refinement \cite{fan2018road}. The performance of these algorithms has been significantly outperformed by end-to-end deep stereo networks, thanks to the recent advancements in deep learning techniques. PSMNet \cite{chang2018psmnet}, GwcNet \cite{guo2019gwcnet}, AANet \cite{xu2020aanet}, LEA-Stereo \cite{cheng2020leastereo}, RAFT-Stereo \cite{lipson2021raftstereo}, and CRE-Stereo \cite{li2022crestereo} are six representative end-to-end deep stereo networks proposed in recent years. PSMNet \cite{chang2018psmnet} employs a spatial pyramid to capture multi-scale information and employs multiple 3D convolutional layers to exploit both local and global contexts for cost computation. GwcNet \cite{guo2019gwcnet}, on the other hand, builds upon the foundation of PSMNet by constructing the cost volume via group-wise correlation, thereby enhancing the 3D stacked hourglass network. In light of the computational demands of 3D convolutions, researchers have actively sought ways to minimize the trade-off between efficiency and accuracy in stereo matching. For example, LEA-Stereo \cite{cheng2020leastereo} introduces the neural architecture search (NAS) \cite{elsken2019neural} technique to stereo matching. This pioneering approach results in the first end-to-end hierarchical NAS framework for deep stereo matching. RAFT-Stereo \cite{lipson2021raftstereo}, a rectified stereo matching method that draws inspiration from the optical flow estimation network RAFT \cite{teed2020raft}, leverages the RAFT architecture to perform accurate and real-time stereo matching inference. The network utilizes recurrent structures to refine correlation features and enhance the disparity estimation accuracy. CRE-Stereo \cite{li2022crestereo}, another recent prior art based on recurrent refinement (to update disparities in a coarse-to-fine manner) and adaptive group correlation (to mitigate the impact of erroneous rectification), yields more compelling disparity estimation results. In this article, we develop our S$^3$M-Net based on the RAFT-Stereo architecture. 
	
	\begin{figure*}[t!]
		\centering
		\includegraphics[width=0.99 \textwidth]{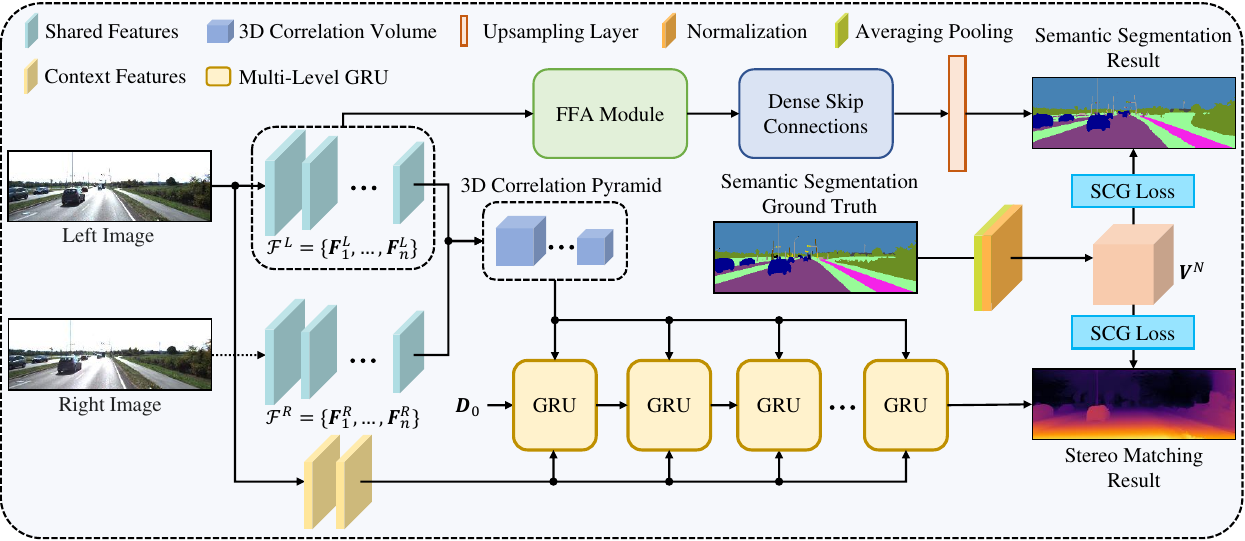}
		\caption{
			The architecture of our proposed S$^3$M-Net for end-to-end joint learning of semantic segmentation and stereo matching.
		}
		\label{fig.pipeline}
	\end{figure*}
	
	\subsection{Multi-Task Joint Learning for Semantic Segmentation and Stereo Matching}
	\label{sec.joint_learning}
	
	Existing frameworks that jointly address semantic segmentation and stereo matching generally focus on improving disparity accuracy by leveraging semantic information \cite{yang2018segstereo, zhang2019dispsegnet, wu2019semantic, dovesi2020real, chen2020sgnet, zhan2019dsnet}, while the discussion regarding the utilization of disparity information to enhance semantic segmentation at the feature level for joint learning remains limited, except for the exploration of ``RGB-X semantic segmentation'' discussed in Sect. \ref{sec.semantic_segmentation}. Nevertheless, these prior arts either require a large amount of well-annotated training data or involve intricate training strategies for the joint learning of both tasks. For instance, SegStereo \cite{yang2018segstereo} and DispSegNet \cite{zhang2019dispsegnet} require an initial unsupervised training phase on the large-scale Cityscapes \cite{cordts2016cityscapes} dataset, followed by a subsequent supervised fine-tuning on the smaller KITTI 2012 and 2015 \cite{geiger2012kitti, menze2015kitti} datasets. Similarly, the studies presented in \cite{wu2019semantic, dovesi2020real, chen2020sgnet} involve the pre-training of their spatial branches (performing stereo matching) on the large-scale SceneFlow \cite{mayer2016sceneflow} dataset, followed by the fine-tuning of both semantic and spatial branches on the KITTI 2012 and 2015 datasets \cite{geiger2012kitti, menze2015kitti}. DSNet \cite{zhan2019dsnet} adopts a different joint learning strategy in which the training alternates between the semantic segmentation and stereo matching networks, with each network being frozen during the training of the other. However, achieving simultaneous convergence of the two networks can be challenging, as the shared features are not learned in an end-to-end manner. Additionally, we were unable to locate publicly available source code (in PyTorch or TensorFlow) for these prior arts, and re-implementations carry the risk of introducing errors. In contrast to the aforementioned approaches, our proposed S$^3$M-Net is trained in an end-to-end fashion and capable of jointly learning semantic segmentation and stereo matching even when the training data are limited.

	\section{Methodology}
	\label{sec.methodology}
	
	As illustrated in Fig. \ref{fig.pipeline}, our proposed S$^3$M-Net consists of five main components:
	\begin{enumerate}[label=(\arabic*)]
		\item Joint encoder to extract shared features from RGB images;
		\item Multi-level GRU update operator to refine disparity maps;
		\item Feature fusion adaptation module to transform shared features into the semantic space and fuse them with features extracted from the disparity maps;
		\item Densely-connected skip connection decoder to decode fused features and produce final semantic predictions;
		\item Semantic consistency-guided loss to supervise the entire joint learning process.
	\end{enumerate}

	\subsection{Joint Encoder}
	\label{sec.joint_encoder}
	Given a pair of well-rectified stereo images $\boldsymbol{I}^L$, $\boldsymbol{I}^R \in \mathbb{R}^{H \times W \times 3}$, where $H$ and $W$ denote their height and width, respectively, we employ a joint encoder consisting of a series of residual blocks and downsampling layers to extract features $\mathcal{F}^L = \left\{\boldsymbol{F}^L_1,\dots,\boldsymbol{F}^L_n \right\}$ and $\mathcal{F}^R = \left\{\boldsymbol{F}^R_1,\dots,\boldsymbol{F}^R_n \right\}$ from $\boldsymbol{I}^L$ and $\boldsymbol{I}^R$, respectively. $\mathcal{F}^L$ is subsequently shared with the semantic segmentation task. 
	
	\subsection{Multi-Level GRU Update Operator}
	\label{sec.disp_decoder}
	{Using the features $\mathcal{F}^L$ and $\mathcal{F}^R$ extracted by the joint encoder, we first construct an initial 3D correlation volume $\boldsymbol{C}_1 \in \mathbb{R}^{H \times W \times W}$ as follows:
		\begin{equation}
			\boldsymbol{C}_1(i,j,k)={\boldsymbol{F}^{L}_{n}}(i,j,:)\cdot{\boldsymbol{F}^{R}_{n}}(i,k,:),
		\end{equation}
		where $i$ represents the $i$-th row, and $j$ and $k$ represent to the $j$-th and $k$-th columns in the left and right shared feature maps, respectively. 
		We then construct a pyramid of 3D correlation volumes $\mathcal{C} = \left\{\boldsymbol{C}_1, \dots, \boldsymbol{C}_m\right\}$ by downsampling $\boldsymbol{C}_1$ with average pooling operations. 
		The $m$-th 3D correlation volume $\boldsymbol{C}_{m} \in \mathbb{R}^{H \times W \times \frac{W}{2^{m-1}}}$ is constructed from the ($m-1$)-th 3D correlation volume $\boldsymbol{C}_{m-1}$} using 1D average pooling with a kernel size of 2 and a stride of 2. 
	Inspired by RAFT-Stereo \cite{lipson2021raftstereo}, we adopt a multi-level GRU update operator to refine a sequence of disparity maps $\mathcal{D} = \left\{ \boldsymbol{D}_1, \dots, \boldsymbol{D}_n \right\} $, where $\boldsymbol{D}_i \in \mathbb{R}^{H \times W}$ ($i=1,\dots,n$). This refinement process is performed in a coarse-to-fine manner, starting from an initial disparity map $\boldsymbol{D}_0$ in which all disparities are initialized to $0$.

	\subsection{Feature Fusion Adaptation Module}
	\label{sec.FFA_module}
	\begin{figure}[t!]
		\centering
		\includegraphics[width=0.49 \textwidth]{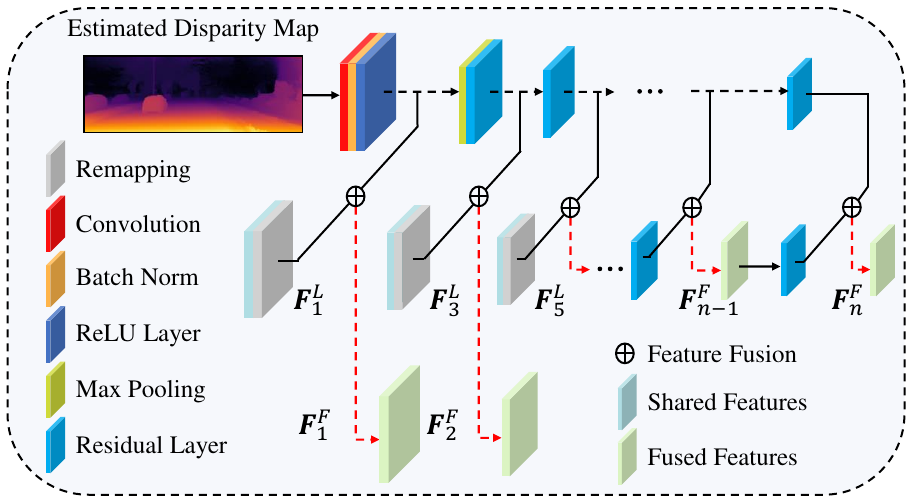}
		\caption{An illustration of our proposed FFA module.
		}
		\label{fig.FFA}
	\end{figure}
	
	In stereo matching, a lower number of channels, \textit{e.g.}, the 256 channels utilized in RAFT-Stereo \cite{lipson2021raftstereo}, is often sufficient for capturing relevant features for 1D correspondence search, especially when considering computational efficiency. On the other hand, semantic segmentation requires pixel-level classification and a more in-depth scene understanding. It benefits from complex feature representations that can capture fine-grained details and object boundaries, making a larger number of channels, \textit{e.g.}, the 2048 channels employed in SNE-RoadSeg \cite{fan2020sneroadseg}, advantageous for this task. Therefore, we introduce the FFA module to align the channels and resolutions between the disparity and semantic feature maps during joint learning.

	As illustrated in Fig \ref{fig.FFA}, given the left shared feature maps $\mathcal{F}^L = \left\{\boldsymbol{F}^L_1,\dots,\boldsymbol{F}^L_n \right\}$ and the disparity map pyramid $\mathcal{D} = \left\{ \boldsymbol{D}_1, \dots, \boldsymbol{D}_n \right\} $, we obtain the adapted fused feature sequence $\mathcal{F}^F = \left\{\boldsymbol{F}^F_1,\dots,\boldsymbol{F}^F_n \right\}$ using our proposed FFA module, which can be formulated as follows: 
	\begin{equation}
		\boldsymbol{F}^F_i = \mathcal{A}_i(\mathcal{F}^L) \oplus \mathcal{E}^D_i(\boldsymbol{D}_n),
	\end{equation}
	where $\mathcal{E}^D$ denotes the disparity map encoding operation, $\oplus$ denotes the feature fusion operation, and $\mathcal{A}_i$ is defined as our feature adaptation operation, as formulated as follows:
	\begin{equation}
		\mathcal{A}_i(\mathcal{F}^L) = \left\{
		\begin{aligned}
			&\mathcal{R}( \boldsymbol{F}^L_{2i-1}), && i \leq \frac{n+1}{2}\\
			&\mathcal{E}( \boldsymbol{F}^{F}_{i-1} \oplus \mathcal{E}^{D}_{i-1}(\boldsymbol{D}_n)), && i > \frac{n+1}{2}
		\end{aligned},
		\right.
	\end{equation} 
	where $\mathcal{R}$ represents the remapping operation from the shared feature space to the semantic feature space, and $\mathcal{E}$ represents the encoding operation for the semantic feature maps. 
	
	Specifically, for the remapping operation $\mathcal{R}$, we employ $3 \times 3$ convolutional layers with a stride of 2 and padding of 1, each followed by a batch normalization layer and a rectified linear unit (ReLU) activation layer, adapting the feature map channels to 64, 256, and 512, respectively. 
	Regarding the disparity encoding operation $\mathcal{E}^D$, we employ ResNet-152 \cite{he2016resnet} as the backbone network to extract features from the last disparity map $\boldsymbol{D}_n$. 
	In ResNet-152, the first block consists of a convolutional layer, a batch normalization layer, and a ReLU activation layer. Then, a max pooling layer and four residual layers are sequentially applied to progressively increase the number of feature map channels.
	
	Similarly, we utilize the residual block for the encoding operation $\mathcal{E}$ on the semantic feature maps, resulting in feature maps with 1024 and 2048 channels. The fused features $\mathcal{F}^F$ contain both texture and spatial geometric information, thereby enhancing semantic scene understanding. 
	We conduct an ablation study for different feature fusion modules in Sect. \ref{sec.exp_ablation}.

	\subsection{Densely-Connected Skip Connection Decoder}
	\label{sec.seg_decoder}
	
	We employ the decoder introduced in our previous work SNE-RoadSeg \cite{fan2020sneroadseg} to decode the fused features and generate the semantic prediction. In this encoder, three convolutional layers in the feature extractor and the upsampling layer share the same parameters: a $3 \times 3$ kernel size, a stride of 1, and a padding of 1.
	In the final layer, features are upsampled to create the prediction map with $N$ channels, where $N$ denotes the number of semantic classes.

	\subsection{Semantic Consistency-Guided Joint Learning Loss }
	\label{sec.scg-loss}
	
	The loss function employed in our joint learning framework should guide the supervision of both the semantic segmentation and stereo matching tasks. 
	Gradient smoothness between the disparity and semantic segmentation maps typically aligns closely, particularly at inter-class boundaries, where traditional training strategies tend to result in more errors due to factors such as occlusion and reflection. In light of this, we propose an SCG loss function to supervise the entire joint learning process, which leverages semantic consistency to optimize the training of S$^3$M-Net.
	
	Given the ground-truth semantic segmentation map $\boldsymbol{M}^{G} \in \mathbb{R}^{H \times W}$, 
	Each pixel $\boldsymbol{p}$ of $\boldsymbol{M}^{G}$ can be written as follows:
	\begin{equation}
		\boldsymbol{M}^{G}(\boldsymbol{p}) \in \{ 1, \dots, C \},
	\end{equation}
	where $C$ refers to the number of the semantic classes. We construct an extended 3D volume ${\boldsymbol{V}^{3D}} \in \mathbb{R}^{H \times W \times C}$ using the following expression: 
	\begin{equation}
		\boldsymbol{V}^{3D}_{c}(\boldsymbol{p}) = \delta (\boldsymbol{M}^{G}(\boldsymbol{p}), c),
	\end{equation} 
	where $c$ represents the $c$-th channel in the volume, and $\delta$ denotes the Kronecker Delta function \cite{okada2019onehot}. As a result, each channel of the volume can be regarded as a binary segmentation map of the $c$-th class. To emphasize semantic consistency, We use an average pooling operation for each channel to obtain the inter-class volume $\boldsymbol{V}^{I}\in \mathbb{R}^{H \times W \times C}$:
	
	\begin{equation}
		\boldsymbol{V}^{I} = \mathcal{P}(\boldsymbol{V}^{3D}),
	\end{equation}
	where $\mathcal{P}$ denotes the average pooling operation. Furthermore, we apply a normalization operation: 
	\begin{equation}
		\boldsymbol{V}^{N}(\boldsymbol{p}) = e^{-(2\boldsymbol{V}^{I}(\boldsymbol{p})-1)^2}.
	\end{equation}
	to obtain a normalized volume $\boldsymbol{V}^{N} \in \mathbb{R}^{H \times W \times C}$.
	We then map $\boldsymbol{V}^{N}$ to a semantic consistency-guided weight map $\boldsymbol{W} \in \mathbb{R}^{H \times W}$ through:
	\begin{equation}
		\boldsymbol{W}(\boldsymbol{p}) = \maxc\limits_{c} \left\{ \boldsymbol{V}_{c}^{N}(\boldsymbol{p}) \right\}.
	\end{equation}
	The total loss function 
	\begin{equation}
		\mathcal{L}_{scg} = \mathcal{L}_{ss} + \mathcal{L}_{sm}
	\end{equation}
	consists of an SCG semantic segmentation loss $\mathcal{L}_{ss}$ and an SCG stereo matching loss $\mathcal{L}_{sm}$. $\mathcal{L}_{ss}$ is formulated as follows: 
	\begin{equation}
		\mathcal{L}_{ss} = -\frac{1}{N} \sum_{i=1}^{N} \sum_{c=1}^{C} [ (1-\alpha) + \alpha \boldsymbol{W}(\boldsymbol{p})] y_{c}(\boldsymbol{p}) \log(\hat{y}_{c}(\boldsymbol{p})),
	\end{equation}
	where $N$ denotes the pixel number, $C$ represents the class number, $\hat{y}_{c}(\boldsymbol{p})$ denotes the predicted probability of $\boldsymbol{p}$ belonging to class $c$, $y_{c}(\boldsymbol{p})$ represents the ground-truth label for $\boldsymbol{p}$ in class $c$, and $\alpha$ denotes the loss weight. Based on the ablation study detailed in Sect. \ref{sec.exp_ablation}, we set the value of $\alpha$ to 0.1. Moreover, $\mathcal{L}_{sm}$ is formulated as follows: 
	\begin{equation}
		\mathcal{L}_{sm}=\sum_{i=1}^{N} [ (1-\alpha) + \alpha \boldsymbol{W}(\boldsymbol{p})] \gamma^{N-i} \left\|\boldsymbol{D}^{G}-\boldsymbol{D}_{i}\right\|_{1},
		\label{eq.scg_loss}
	\end{equation}
	where $\boldsymbol{D}^{G}$ represents the ground-truth disparity map and $\boldsymbol{D}_{i}$ denotes the $i$-th disparity map in $\mathcal{D}$. 
	$\alpha$ is set to 0.1 and $\gamma$ is set to 0.9. 
	
	\begin{table*}[t!]
		\fontsize{6.8}{11.5}\selectfont
		\centering
		\caption{
			Comparisons of SoTA semantic segmentation networks on the vKITTI2 \cite{cabon2020vkitti2} dataset. The symbol $\uparrow$ indicates that a higher value corresponds to better performance. The best results are shown in bold font.
		}
		\label{tab.seg_vkitti}
		\setlength{\tabcolsep}{1.5mm}
		\begin{tabular}
			{C{1cm}| L{2.6cm} |C{1.45cm} C{1.45cm} C{1.45cm} C{1.45cm} C{1.45cm} C{1.45cm} C{1.45cm}}
			\toprule
			Category & \multicolumn{1}{c|}{Networks} & Acc (\%) $\uparrow$ & mAcc (\%) $\uparrow$ & mIoU (\%) $\uparrow$ & fwIoU (\%) $\uparrow$ & Pre (\%) $\uparrow$ & Rec (\%) $\uparrow$ & FSc (\%) $\uparrow$ \\
			\hline
			\hline
			\multirow{10}{*}{\rotatebox{90}{Single-Modal}}
			& SegNet \cite{badrinarayanan2017segnet} & 59.29 & 32.54 & 23.93 & 48.17 & 66.10 & 66.73 & 61.11 \\
			& U-Net \cite{ronneberger2015unet} & 62.71 & 37.65 & 29.83 & 55.10 & 75.80 & 67.67 & 65.26 \\
			& PSPNet \cite{zhao2017pspnet} & 76.26 & 53.53 & 44.81 & 69.30 & 81.55 & 79.68 & 75.38 \\
			& DeepLabv3+ \cite{chen2018deeplabv3plus} & 92.19 & 63.15 & 56.90 & 87.15 & 89.00 & 92.71 & 90.16 \\
			& HRNet \cite{sun2019hrnet} & 74.79 & 40.82 & 32.47 & 63.23 & 73.69 & 76.50 & 73.39 \\
			& BiSeNet V2 \cite{yu2021bisenetv2} & 81.77 & 51.07 & 44.45 & 74.71 & 83.23 & 82.19 & 80.67 \\
			& Segmenter \cite{strudel2021segmenter} & 90.39 & 60.33 & 52.99 & 83.47 & 88.05 & 87.89 & 87.70 \\
			& SegFormer \cite{xie2021segformer} & 94.75 & 70.56 & 64.98 & 90.49 & 93.57 & 93.62 & 93.46 \\
			& Mask2Former \cite{cheng2022mask2former} & 89.29 & 64.58 & 57.14 & 83.84 & 90.75 & 87.23 & 87.19 \\
			& DDRNet \cite{hong2021ddrnet} & 70.80 & 40.32 & 32.10 & 61.44 & 76.35 & 71.67 & 70.57 \\
			\hline
			\multirow{8}{*}{\rotatebox{90}{Feature-Fusion}}
			& FuseNet \cite{hazirbas2017FuseNet} & 49.42 & 31.21 & 22.56 & 41.07 & 79.39 & 50.67 & 47.50 \\
			& MFNet \cite{ha2017MFNet} & 76.22 & 51.50 & 43.41 & 68.82 & 82.46 & 78.65 & 73.80 \\
			& RTFNet \cite{sun2019RTFNet} & 85.22 & 49.47 & 42.59 & 77.69 & 83.74 & 89.17 & 84.41 \\
			& SNE-RoadSeg \cite{fan2020sneroadseg} & 83.64 & 60.85 & 52.56 & 75.14 & 83.44 & 81.66 & 77.77 \\
			& OFF-Net \cite{min2022orfd} & 90.84 & 61.51 & 55.27 & 84.69 & 89.24 & 86.71 & 86.15\\
			& RoadFormer \cite{li2023roadformer} & 97.54 & 86.58 & 80.83 & 95.34 & 96.99 & 96.86 & 96.91 \\
			\cline{2-9}
			& \textbf{S$^3$M-Net}  & 98.27 & \textbf{88.28} & \textbf{84.25} & 96.92 & 98.29 & \textbf{98.32} & 98.28 \\
			& \textbf{S$^3$M-Net w/ SCG loss}  & \textbf{98.32} & 88.24 & 84.18 & \textbf{96.98} & \textbf{98.37} & 98.28 & \textbf{98.31} \\
			\bottomrule
		\end{tabular}
	\end{table*}
	
	\begin{table*}[t!]
		\fontsize{6.8}{11.5}\selectfont
		\centering
		\caption{
			Comparisons of SoTA semantic segmentation networks on the KITTI 2015 \cite{menze2015kitti} dataset. The symbol $\uparrow$ indicates that a higher value corresponds to better performance. The best results are shown in bold font.
		}
		\label{tab.seg_kitti}
		\setlength{\tabcolsep}{1.5mm}
		\begin{tabular}
			{C{1cm}| L{2.4cm} |C{1.45cm} C{1.45cm} C{1.45cm} C{1.45cm} C{1.45cm} C{1.45cm} C{1.45cm}}
			\toprule
			Category & \multicolumn{1}{c|}{Networks} & Acc (\%) $\uparrow$ & mAcc (\%) $\uparrow$ & mIoU (\%) $\uparrow$ & fwIoU (\%) $\uparrow$ & Pre (\%) $\uparrow$ & Rec (\%) $\uparrow$ & FSc (\%) $\uparrow$ \\
			\hline
			\hline
			\multirow{10}{*}{\rotatebox{90}{Single-Modal}}
			& SegNet \cite{badrinarayanan2017segnet} & 59.63 & 31.98 & 22.61 & 43.98 & 55.25 & 67.49 & 57.29 \\
			& U-Net \cite{ronneberger2015unet} & 69.02 & 41.15 & 30.64 & 55.65 & 69.11 & 77.65 & 71.04 \\
			& PSPNet \cite{zhao2017pspnet} & 80.03 & 44.97 & 38.15 & 68.62 & 79.29 & 82.66 & 79.59 \\
			& DeepLabv3+ \cite{chen2018deeplabv3plus} & 82.33 & 50.15 & 42.79 & 72.43 & 83.85 & 87.18 & 84.59 \\
			& HRNet \cite{sun2019hrnet} & 63.42 & 31.68 & 22.78 & 49.40 & 55.10 & 67.71 & 57.21 \\
			& BiSeNet V2 \cite{yu2021bisenetv2} & 73.68 & 41.66 & 32.71 & 60.55 & 68.35 & 81.79 & 72.37 \\
			& Segmenter \cite{strudel2021segmenter} & 84.53 & 50.77 & 43.63 & 74.72 & 82.99 & 87.15 & 84.41 \\
			& SegFormer \cite{xie2021segformer} & 88.28 & 59.23 & 51.39 & 80.53 & 87.15 & 90.85 & 88.46 \\
			& Mask2Former \cite{cheng2022mask2former} & 84.35 & 54.33 & 45.87 & 75.56 & 84.74 & 89.12 & 85.92 \\
			& DDRNet \cite{hong2021ddrnet} & 62.12 & 31.61 & 22.63 & 48.15 & 57.09 & 68.98 & 59.07 \\
			\hline
			\multirow{8}{*}{\rotatebox{90}{Feature-Fusion}}
			& FuseNet \cite{hazirbas2017FuseNet} & 41.79 & 19.05 & 11.38 & 27.53 & 44.14 & 44.35 & 37.68 \\
			& MFNet \cite{ha2017MFNet} & 81.02 & 48.13 & 40.70 & 70.42 & 82.85 & 85.73 & 82.36 \\
			& RTFNet \cite{sun2019RTFNet} & 71.61 & 39.26 & 30.35 & 57.98 & 69.52 & 85.16 & 74.28 \\
			& SNE-RoadSeg \cite{fan2020sneroadseg} & 79.46 & 51.91 & 41.56 & 69.22 & 81.45 & 87.05 & 82.91 \\
			& OFF-Net \cite{min2022orfd} & 75.84 & 40.13 & 33.13 & 64.02 & 77.48 & 72.19 & 70.62 \\
			& RoadFormer \cite{li2023roadformer} & 90.05 & 62.34 & 55.13 & 83.40 & \textbf{91.65} & 91.39 & 91.11 \\
			\cline{2-9}
			& \textbf{S$^3$M-Net}  & 90.01 & 62.48 & 54.33 & 83.44 & 88.96 & 93.52 & 90.65 \\
			& \textbf{S$^3$M-Net w/ SCG loss} & \textbf{90.66} & \textbf{65.90} & \textbf{57.80} & \textbf{84.53} & 90.85 & \textbf{93.55} & \textbf{91.80} \\
			\bottomrule
		\end{tabular}
	\end{table*}
	
	\begin{figure*}[t!]
		\centering
		\includegraphics[width=0.99 \textwidth]{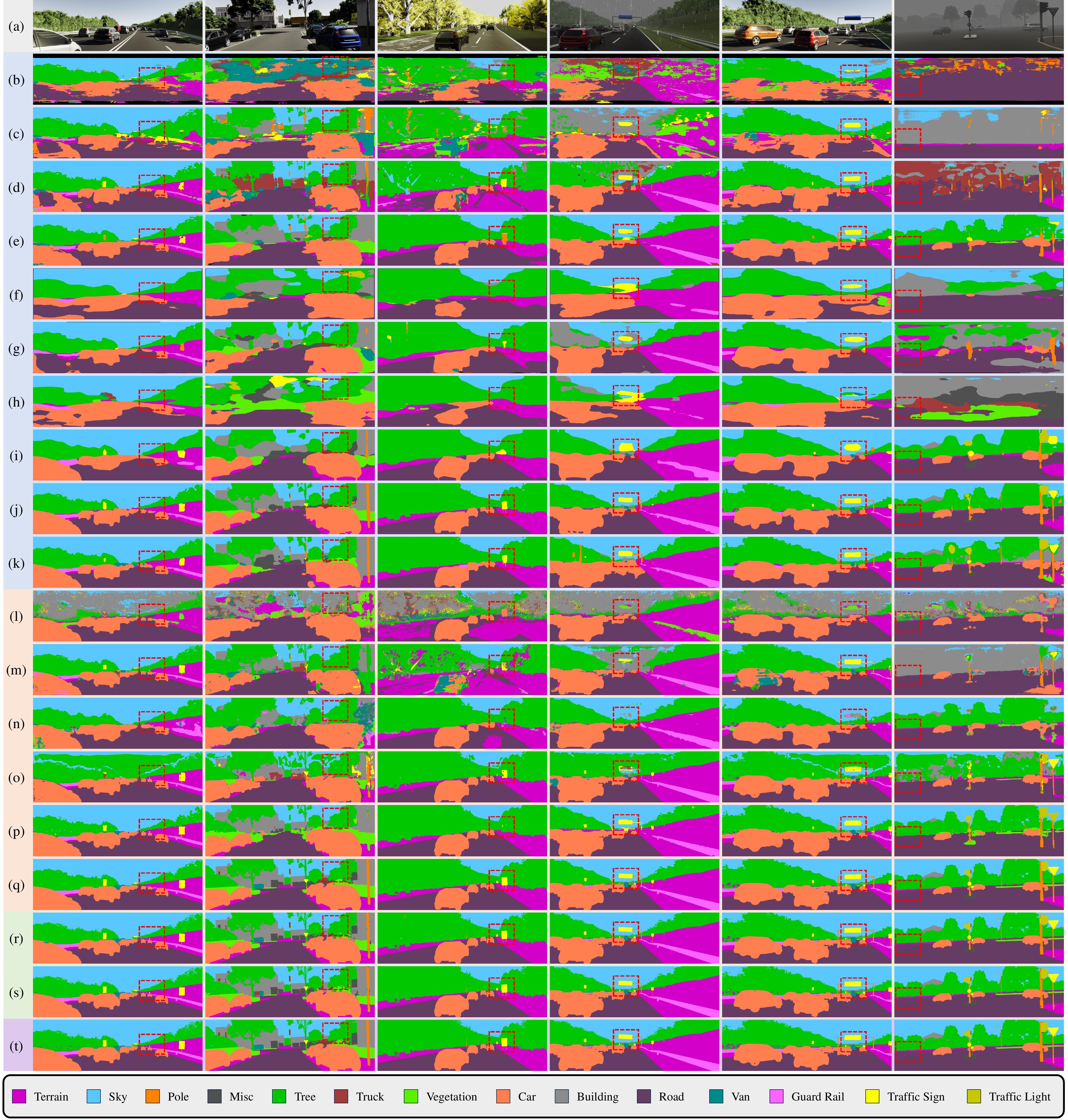}
		\caption{Qualitative experimental results of semantic segmentation on the vKITTI2 \cite{cabon2020vkitti2} dataset: (a) RGB images; (b)-(k) semantic segmentation results achieved by SegNet \cite{badrinarayanan2017segnet}, U-Net \cite{ronneberger2015unet}, PSPNet \cite{zhao2017pspnet}, DeepLabv3+ \cite{chen2018deeplabv3plus}, HRNet \cite{sun2019hrnet}, BiSeNet V2 \cite{yu2021bisenetv2}, Segmenter \cite{strudel2021segmenter}, SegFormer \cite{xie2021segformer}, Mask2Former \cite{cheng2022mask2former}, and DDRNet \cite{hong2021ddrnet}, respectively; (l)-(q) semantic segmentation results achieved by FuseNet \cite{hazirbas2017FuseNet}, MFNet \cite{ha2017MFNet}, RTFNet \cite{sun2019RTFNet}, SNE-RoadSeg \cite{fan2020sneroadseg}, OFF-Net \cite{min2022orfd}, and RoadFormer \cite{li2023roadformer}, respectively; (r)-(s) semantic segmentation results achieved by our proposed S$^3$M-Net w/o and w/ the use of the SCG loss, respectively; (t) ground truth annotations.}
		\label{fig.seg_vkitti}
	\end{figure*}
	
	\begin{figure*}[t!]
		\centering
		\includegraphics[width=0.99 \textwidth]{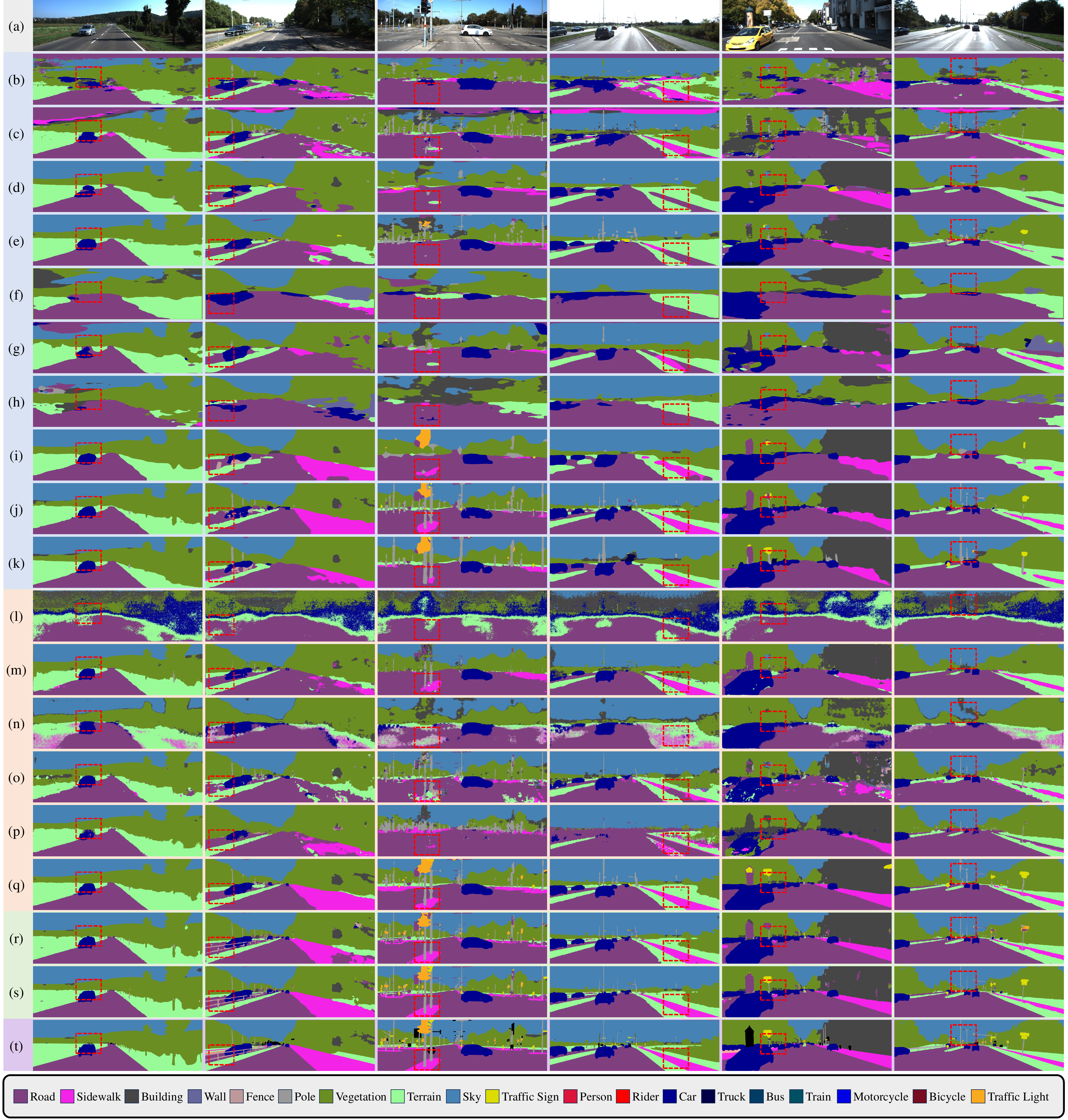}
		\caption{
			Qualitative experimental results of semantic segmentation on the KITTI 2015 \cite{menze2015kitti} dataset: (a) RGB images; (b)-(k) semantic segmentation results achieved by SegNet \cite{badrinarayanan2017segnet}, U-Net \cite{ronneberger2015unet}, PSPNet \cite{zhao2017pspnet}, DeepLabv3+ \cite{chen2018deeplabv3plus}, HRNet \cite{sun2019hrnet}, BiSeNet V2 \cite{yu2021bisenetv2}, Segmenter \cite{strudel2021segmenter}, SegFormer \cite{xie2021segformer}, Mask2Former \cite{cheng2022mask2former}, and DDRNet \cite{hong2021ddrnet}, respectively; (l)-(q) semantic segmentation results achieved by FuseNet \cite{hazirbas2017FuseNet}, MFNet \cite{ha2017MFNet}, RTFNet \cite{sun2019RTFNet}, SNE-RoadSeg \cite{fan2020sneroadseg}, OFF-Net \cite{min2022orfd}, and RoadFormer \cite{li2023roadformer}, respectively; (r)-(s) semantic segmentation results achieved by our proposed S$^3$M-Net w/o and w/ the use of the SCG loss, respectively; (t) ground truth annotations.}
		\label{fig.seg_kitti}
	\end{figure*}
	
	\begin{figure*}[t!]
		\centering
		\includegraphics[width=0.99 \textwidth]{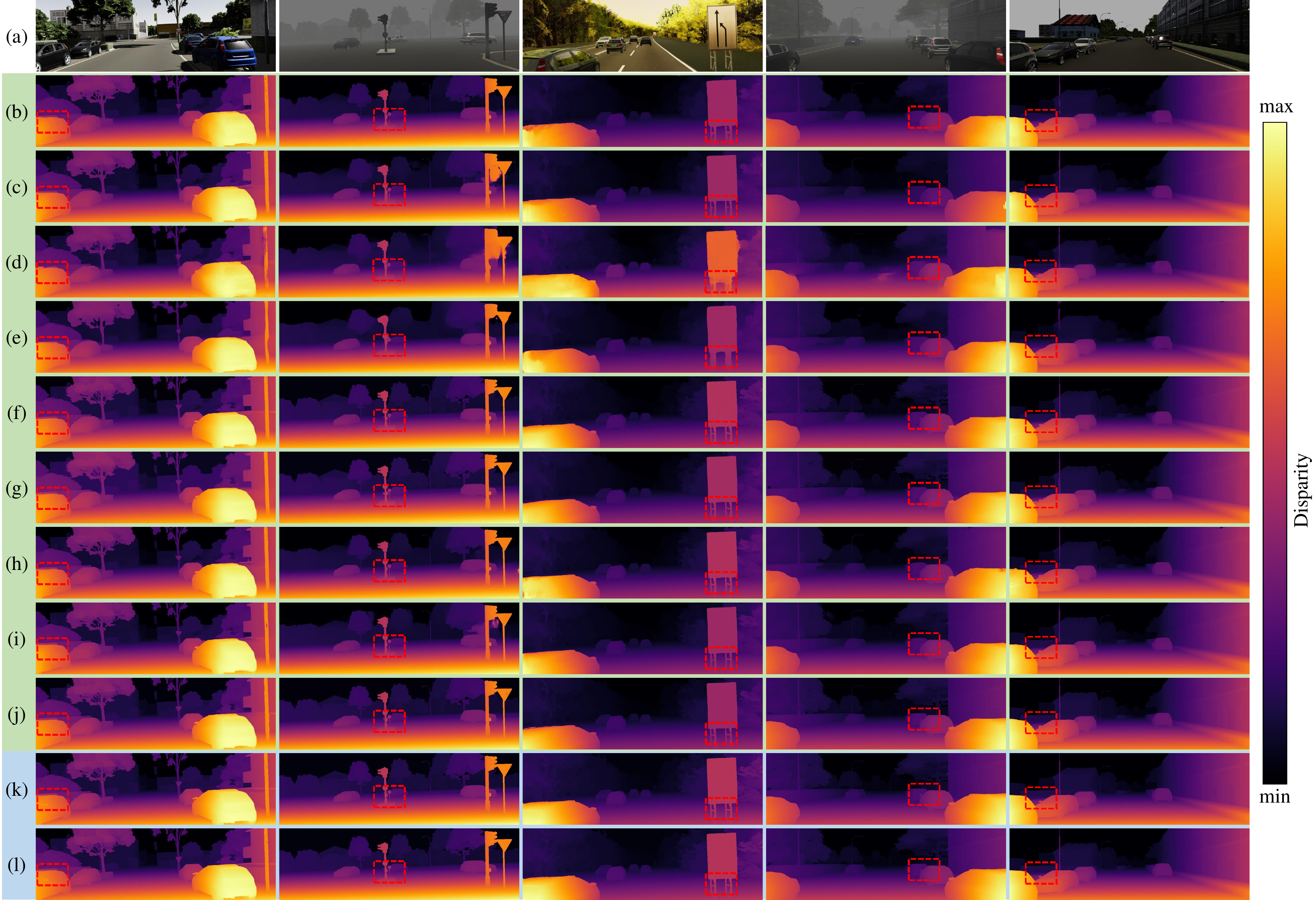}
		\caption{Qualitative experimental results of stereo matching on the vKITTI2 \cite{cabon2020vkitti2} dataset: (a) left RGB images; (b)-(j) disparity maps estimated using PSMNet \cite{chang2018psmnet}, GwcNet \cite{guo2019gwcnet}, AANet \cite{xu2020aanet}, LEA-Stereo \cite{cheng2020leastereo}, RAFT-Stereo \cite{lipson2021raftstereo}, CRE-Stereo \cite{li2022crestereo}, ACVNet \cite{xu2022acvnet}, PCWNet\cite{shen2022pcwnet}, and IGEV-Stereo \cite{xu2023igevstereo}, respectively; (k)-(l) disparity maps estimated using our proposed S$^3$M-Net w/o and w/ the use of the SCG loss, respectively.
		}
		\label{fig.disp_vkitti}
	\end{figure*}
	
	\begin{figure*}[t!]
		\centering
		\includegraphics[width=0.99 \textwidth]{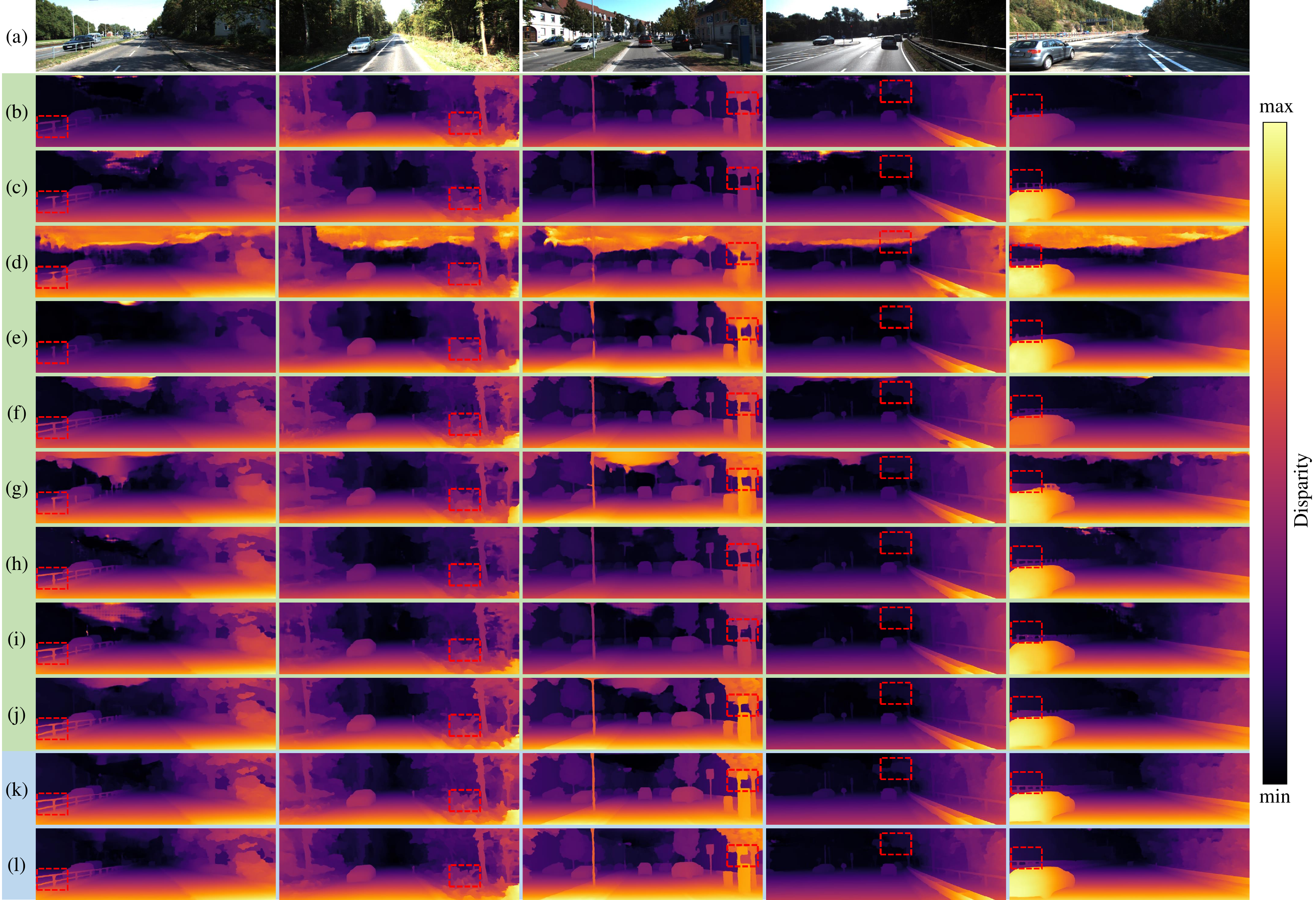}
		\caption{
			Qualitative experimental results of stereo matching on the KITTI 2015 \cite{menze2015kitti} dataset: (a) left RGB images; (b)-(j) disparity maps estimated using PSMNet \cite{chang2018psmnet}, GwcNet \cite{guo2019gwcnet}, AANet \cite{xu2020aanet}, LEA-Stereo \cite{cheng2020leastereo}, RAFT-Stereo \cite{lipson2021raftstereo}, CRE-Stereo \cite{li2022crestereo}, ACVNet \cite{xu2022acvnet}, PCWNet\cite{shen2022pcwnet}, and IGEV-Stereo \cite{xu2023igevstereo}, respectively; (k)-(l) disparity maps estimated using our proposed S$^3$M-Net w/o and w/ the use of the SCG loss, respectively.
		}
		\label{fig.disp_kitti}
	\end{figure*}
	
	\begin{figure}[t!]
		\centering
		\includegraphics[width=0.49 \textwidth]{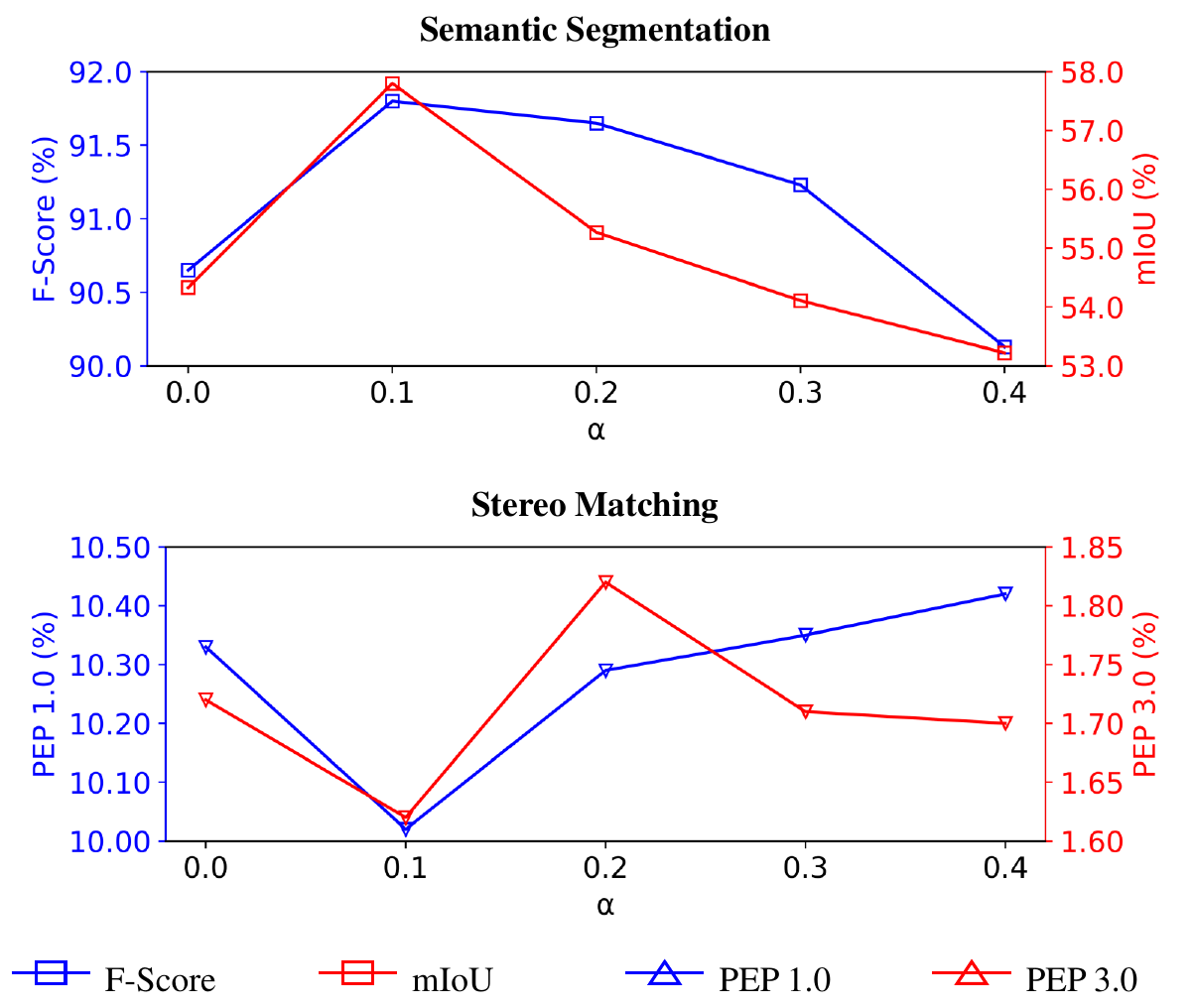}
		\caption{
			Ablation study on the selection of $\alpha$ in the SCG joint learning loss function on the KITTI 2015 \cite{menze2015kitti} dataset.
		}
		\label{fig.ablation}
	\end{figure}

	\section{Experiments}
	\label{sec.experiments}
	In this article, we conduct extensive experiments to evaluate the performance of our developed S$^3$M-Net both quantitatively and qualitatively. The following subsections provide details on the used datasets, experimental set-up, evaluation metrics, and the comprehensive evaluation of our proposed method.
	
	\subsection{Datasets}
	\label{sec.datasets.vkitti}
	Since the training of our network requires both semantic and disparity annotations, we employ two public datasets to evaluate its performance: the vKITTI2 \cite{cabon2020vkitti2} dataset (synthetic yet large-scale) and the KITTI 2015 \cite{menze2015kitti} dataset (real-world yet modest-scale). Their details are as follows:
	\begin{itemize}
		\item 
		The vKITTI2 dataset contains virtual replicas of five sequences from the KITTI dataset. It provides 15 classes for the semantic segmentation tasks. Dense ground-truth disparity maps are acquired through depth rendering using a virtual engine. In our experiments, we randomly select 700 pairs of stereo images, along with their semantic and disparity annotations to evaluate the effectiveness and robustness of our proposed S$^3$M-Net, where 500 pairs are used for model training and the remaining 200 pairs are used for model validation.
		
		\item The KITTI 2015 dataset contains 400 pairs of stereo images captured in real-world driving scenarios, with 200 pairs containing ground truth and the other 200 pairs lacking ground truth. It provides 19 classes for the semantic segmentation tasks (in alignment with the Cityscapes \cite{cordts2016cityscapes} dataset). Sparse disparity ground truth is obtained using a Velodyne HDL-64E LiDAR. In our experiments, we allocate 70\% of the dataset for training, while the remaining portion is used as the test set.
	\end{itemize}
	
	\subsection{Experimental Setup}
	\label{sec.experimental_setup}
	
	Our experiments are conducted on an NVIDIA RTX 3090 GPU. The batch size is set to 1. The maximum disparity search range is set to 192 pixels. All images are cropped to 1000 $\times$ 320 pixels before feeding into the network. We utilize the AdamW \cite{loshchilov2017adamw} optimizer for model training, setting the epsilon and weight decay parameters to $10^{-8}$ and $10^{-5}$, respectively. The initial learning rate is set to $2\times 10^{-4}$. Training lasts for 100K iterations on the vKITTI2 dataset and 20K iterations on the KITTI 2015 dataset. We employ traditional data augmentation techniques to enhance the robustness of the models.
	
	\subsection{Evaluation Metrics}
	\label{sec.evaluation_metrics}
	
	Since our proposed S$^3$M-Net simultaneously performs semantic segmentation and stereo matching, we evaluate the performance of both tasks in our experiments. 
	
	We utilize seven evaluation metrics to quantify the performance of semantic segmentation: (1) accuracy (Acc), (2) mean accuracy (mAcc), (3) mean intersection over union (mIoU), (4) frequency-weighted intersection over union (fwIoU) \cite{zhao2017fwiou}, (5) precision (Pre), (6) recall (Rec), and (7) F1-score (FSc).
	
	Additionally, we use two evaluation metrics: (1) the average end-point error (EPE) and (2) the percentage of error pixels (PEP), setting the tolerance for the latter to 1.0 and 3.0 pixels, respectively, to quantify the performance of stereo matching.

	\subsection{Semantic Segmentation Performance}
	\label{sec.exp_seg}
	
	The qualitative experimental results on the vKITTI2 and KITTI datasets are presented in Figs. \ref{fig.seg_vkitti} and \ref{fig.seg_kitti}, respectively, while the quantitative experimental results on the vKITTI2 and KITTI datasets are given in Tables \ref{tab.seg_vkitti} and \ref{tab.seg_kitti}, respectively. These results suggest that S$^3$M-Net outperforms other SoTA single-modal and feature-fusion networks (including CNN-based and Transformer-based methods) across all evaluation metrics on both datasets. Specifically, it is noteworthy that when the entire joint learning framework is trained by minimizing our proposed SCG loss, S$^3$M-Net achieves the best performance on the KITTI dataset across all evaluation metrics except for Pre. Compared with SoTA methods, it shows improvements of $5.71\%$ in mAcc, $4.84\%$ in mIoU, $1.35\%$ in fwIoU, and $0.76\%$ in FSc, respectively. Similarly, it outperforms other networks on the vKITTI2 dataset in most evaluation metrics, with improvements of $1.72\%$ in fwIoU and $1.44\%$ in FSc. However, for mAcc, mIoU, and Rec, its performance is comparable to that of S$^3$M-Net trained without using the SCG loss. Additionally, it is obvious that feature-fusion networks consistently outperform single-modal networks, particularly under challenging weather and lighting conditions. This observation aligns with our expectations, as feature-fusion networks leverage both RGB images and disparity maps, allowing them to effectively learn informative spatial geometric representations. However, SoTA feature-fusion networks may exhibit higher error rates in distant regions. For instance, FuseNet and SNE-RoadSeg demonstrate poor performance in the sky. We attribute this phenomenon to the deep structure of the encoders, where distinguishing distant objects using disparity features becomes challenging, and the feature fusion process amplifies the influence of the disparity feature. In contrast, within our proposed joint learning framework, we can extract more informative features benefiting from both tasks, irrespective of dataset size. This improvement is likely due to the fact that joint learning of multiple interconnected tasks introduces a form of regularization, which has shown its superiority over uniform complexity penalization in reducing over-fitting.
	
	\begin{table}[t!]
		\fontsize{6.8}{11.5}\selectfont
		\centering
		\caption{
			Comparisons of SoTA stereo matching network on the vKITTI2 \cite{cabon2020vkitti2} dataset. The symbol $\downarrow$ indicates that a lower value corresponds to better performance. The best results are shown in bold font.
		}
		\label{tab.disp_vkitti}
		\setlength{\tabcolsep}{1.5mm}
		\begin{threeparttable}
			\begin{tabular}
				{L{2.4cm}|C{1.6cm} C{1.45cm} C{1.45cm}}
				\toprule
				\multicolumn{1}{c|}{\multirow{2}{*}{Networks}} & \multirow{2}{*}{EPE (pixels) $\downarrow$} & \multicolumn{2}{c}{PEP (\%) $\downarrow$} \\
				\cline{3-4}
				& & $>1$ pixel & $>3$ pixels\\
				\hline   \hline
				PSMNet \cite{chang2018psmnet} & 0.68 & 10.31 & 3.77 \\
				GwcNet \cite{guo2019gwcnet} & 0.65 & 9.72 & 3.69 \\
				AANet \cite{xu2020aanet} & 1.36 & 15.61 & 6.98 \\
				LEA-Stereo \cite{cheng2020leastereo} & 0.83 & 13.33 & 4.84 \\
				RAFT-Stereo \cite{lipson2021raftstereo} & 0.40 & 5.88 & 2.67 \\
				CRE-Stereo \cite{li2022crestereo} & 0.63 & 10.35 & 3.90 \\
				ACVNet \cite{xu2022acvnet} & 0.61 & 9.41 & 3.45 \\
				PCW-Net \cite{shen2022pcwnet} & 0.63 & 9.45 & 3.49 \\
				IGEV-Stereo \cite{xu2023igevstereo} & 0.47 & 7.15 & 3.09 \\
				\hline
				\textbf{S$^3$M-Net}  & 0.39 & 5.59 & 2.55 \\
				\textbf{S$^3$M-Net w/ SCG loss} & \textbf{0.38} & \textbf{5.56} & \textbf{2.55} \\
				\Xhline{1.5pt}
			\end{tabular}
		\end{threeparttable}
	\end{table}
	
	\begin{table}[t!]
		\fontsize{6.8}{11.5}\selectfont
		\centering
		\caption{
			Comparisons of SoTA stereo matching network on the KITTI 2015 \cite{menze2015kitti} dataset. The symbol $\downarrow$ indicates that a lower value corresponds to better performance. The best results are shown in bold font.
		}
		\label{tab.disp_kitti}
		\setlength{\tabcolsep}{1.5mm}
		\begin{threeparttable}
			\begin{tabular}
				{L{2.4cm}|C{1.6cm} C{1.45cm} C{1.45cm}}
				\toprule
				\multicolumn{1}{c|}{\multirow{2}{*}{Networks}} & \multirow{2}{*}{EPE (pixels) $\downarrow$} & \multicolumn{2}{c}{PEP (\%) $\downarrow$} \\
				\cline{3-4}
				& & $>1$ pixel & $>3$ pixels\\
				\hline   \hline
				PSMNet \cite{chang2018psmnet} & 0.74 & 16.12 & 2.61 \\
				GwcNet \cite{guo2019gwcnet} & 0.68 & 14.21 & 2.01 \\
				AANet \cite{xu2020aanet} & 1.10 & 22.67 & 5.37 \\
				LEA-Stereo \cite{cheng2020leastereo} & 0.83 & 18.67 & 3.22 \\
				RAFT-Stereo \cite{lipson2021raftstereo} & 0.60 & 10.78 & 1.96 \\
				CRE-Stereo \cite{li2022crestereo} & 0.92 & 19.68 & 3.35 \\
				ACVNet \cite{xu2022acvnet} & 0.68 & 13.93 & 2.10 \\
				PCW-Net \cite{shen2022pcwnet} & 0.70 & 14.81 & 2.43 \\
				IGEV-Stereo \cite{xu2023igevstereo} & 0.62 & 12.15 & 1.99 \\
				\hline
				\textbf{S$^3$M-Net}  & 0.56 & 10.33 & 1.72 \\
				\textbf{S$^3$M-Net w/ SCG loss} & \textbf{0.55} & \textbf{10.02} & \textbf{1.62} \\
				\Xhline{1.5pt}
			\end{tabular}
		\end{threeparttable}
	\end{table}
	
	\subsection{Stereo Matching Performance} 
	\label{sec.exp_disp}
	
	The qualitative experimental results on the vKITTI2 and KITTI datasets are given in Figs. \ref{fig.disp_vkitti} and \ref{fig.disp_kitti}, respectively, while the quantitative experimental results on the vKITTI2 and KITTI datasets are presented in Tables \ref{tab.disp_vkitti} and \ref{tab.disp_kitti}, respectively. These results suggest that S$^3$M-Net outperforms other SoTA stereo matching networks across all evaluation metrics on both datasets. Specifically, S$^3$M-Net trained with and without using the SCG loss achieves the top and second-best overall performances, respectively. S$^3$M-Net, when trained without using the SCG loss, demonstrates improvements of $2.50\%$-$71.32\%$ in EPE, $4.17\%$-$64.19\%$ in PEP 1.0, and $4.49\%$-$67.97\%$ in PEP 3.0. On the other hand, S$^3$M-Net, when trained with the SCG loss, shows improvements of $5.00\%$-$72.06\%$ in EPE, $5.44\%$-$64.38\%$ in PEP 1.0, and $4.49\%$-$69.83\%$ in PEP 3.0. We attribute these improvements to the feature sharing and fusion strategies applied in S$^3$M-Net. First, sharing features with the semantic segmentation task allows S$^3$M-Net to learn stereo matching effectively even with limited training data. 
	Second, as discussed above, stereo matching can sometimes produce ambiguous disparity estimations, especially in occluded or texture-less areas. The pursuit of semantic consistency helps resolve such ambiguities, leading to more reliable disparity estimation results.
	In Fig. \ref{fig.disp_kitti}, it is evident that regions lacking disparity ground truth frequently have substantial errors. Previous stereo matching algorithms have endeavored to tackle this issue through knowledge distillation with pre-trained models \cite{xu2020aanet}. Nevertheless, our S$^3$M-Net successfully overcomes this challenge by leveraging semantic information.
	
	\begin{table*}[t!]
		\fontsize{6.8}{11.5}\selectfont
		\centering
		\caption{
			Ablation study on feature fusion strategy in our FFA module on the KITTI \cite{menze2015kitti} 2015 dataset.
		}
		\label{tab.ablation_fusion}
		\setlength{\tabcolsep}{1.5mm}
		\begin{tabular}
			{L{1.7cm}| C{1.45cm} C{1.45cm} C{1.45cm} C{1.45cm} C{1.45cm} C{1.45cm} C{1.45cm}}
			\toprule
			Fusion Strategy & Acc (\%) $\uparrow$ & mAcc (\%) $\uparrow$ & mIoU (\%) $\uparrow$ & fwIoU (\%) $\uparrow$ & Pre (\%) $\uparrow$ & Rec (\%) $\uparrow$ & FSc (\%) $\uparrow$ \\
			\hline
			\hline
			Addition & \textbf{90.01} & \textbf{62.48} & \textbf{54.33} & \textbf{83.44} & \textbf{88.96} & 93.52 & \textbf{90.65} \\
			Concatenation & 86.88 & 57.43 & 48.40 & 78.50 & 85.96 & \textbf{93.71} & 88.92 \\
			CFM \cite{wei2020cfm} & 86.87 & 57.41 & 48.77 & 79.13 & 85.41 & 92.05 & 87.63 \\
			DDPM \cite{pang2020ddpm} & 86.52 & 58.65 & 49.51 & 78.14 & 85.56 & 93.34 & 88.44 \\
			SA Gate \cite{chen2020sagate} & 87.62 & 61.77 & 52.10 & 80.55 & 86.74 & 92.31 & 88.52 \\
			SWS \cite{wang2020AysmFusion, wang2020CEN, wang2022TokenFusion} & 87.94 & 58.11 & 49.64 & 80.25 & 86.63 & 93.34 & 89.20 \\
			\bottomrule
		\end{tabular}
	\end{table*}
	
	\subsection{Ablation studies}
	\label{sec.exp_ablation}
	
	In this subsection, we first conduct an ablation study on the selection of loss weight $\alpha$ in (\ref{eq.scg_loss}). Fig. \ref{fig.ablation} shows the quantitative experimental results with respect to different $\alpha$ in the range of $0.0$ to $0.4$ for both semantic segmentation and stereo matching. It can be obvious that when $\alpha=0.1$, S$^3$M-Net achieves the best overall performance for both tasks. Further weight tuning is possible, but it should be approached cautiously, especially when dealing with limited data to avoid over-fitting.
	
	Furthermore, we conduct an additional ablation study on the feature fusion strategy in our proposed FFA module. As shown in Table \ref{tab.ablation_fusion}, when using the addition operation to fuse heterogeneous features, the FFA module consistently achieves the best performance across all evaluation metrics, compared to other feature fusion strategies, including concatenation, cross feature module (CFM) \cite{wei2020cfm}, dynamic dilated pyramid module (DDPM) \cite{pang2020ddpm}, separation-and-aggregation gate (SA Gate) \cite{chen2020sagate}, and softmax weighted sum (SWS) used in AysmFusion \cite{wang2020AysmFusion}, CEN \cite{wang2020CEN}, and TokenFusion \cite{wang2022TokenFusion}. 
	
	\section{Discussion}
	\label{sec.discussion}
	
	The experimental results shown in Sect. \ref{sec.experiments} provide strong support for the claims made in Sect. \ref{sec.intro}. First, the joint learning of semantic segmentation and stereo matching, two interconnected environmental perception tasks, using our proposed S$^3$M-Net introduces a form of regularization that has shown its effectiveness in reducing overfitting, particularly in scenarios where training data are limited. Secondly, this end-to-end joint learning framework yields improved performance when compared to the models trained separately for each task. Finally, the pursuit of semantic consistency in joint learning helps reduce ambiguous disparity estimations in texture-less or occluded regions. We believe that our proposed S$^3$M-Net can be readily deployed in autonomous vehicles after addressing the following limitations:
	\begin{itemize}
		\item S$^3$M-Net requires both semantic and disparity annotations, and collecting data with such ground truth remains a labor-intensive process. Therefore, the exploration of potential solutions such as semi-supervised or low/few-shot semantic segmentation and un/self-supervised stereo matching is a promising avenue for future research.
		\item S$^3$M-Net achieves a processing speed of 0.66 fps when processing input RGB images with a resolution of 1248 $\times$ 384 pixels. We believe that further computational efficiency optimizations are necessary before deploying S$^3$M-Net in autonomous vehicles.
	\end{itemize}
	
	\section{Conclusion}
	\label{sec.conclusion}
	
	This article introduced S$^3$M-Net, an effective solution for joint learning of semantic segmentation and stereo matching. We have made three significant contributions in this work: (1) the development of an entire joint learning framework that shares features between both tasks and fuses heterogeneous features to improve semantic segmentation, (2) a feature fusion adaption module designed to enable effective feature sharing between the two tasks, and (3) a semantic consistency-guided joint learning loss that emphasizes structural consistency in both tasks. We conducted extensive experiments on the vKITTI2 (synthetic and large) and KITTI (real-world and small) datasets to validate the effectiveness of our framework, the FFA module, and the training loss. Our results demonstrate the superior performance of our approach compared to all other existing methods. 
	
	\normalem
	\bibliographystyle{IEEEtran}


\end{document}

%% file: packages.tex
\usepackage[table,xcdraw]{xcolor}
\usepackage{scalerel}
\usepackage{tikz}
\usetikzlibrary{svg.path}

\definecolor{orcidlogocol}{HTML}{A6CE39}
\tikzset{
  orcidlogo/.pic={
    \fill[orcidlogocol] svg{M256,128c0,70.7-57.3,128-128,128C57.3,256,0,198.7,0,128C0,57.3,57.3,0,128,0C198.7,0,256,57.3,256,128z};
    \fill[white] svg{M86.3,186.2H70.9V79.1h15.4v48.4V186.2z}
                 svg{M108.9,79.1h41.6c39.6,0,57,28.3,57,53.6c0,27.5-21.5,53.6-56.8,53.6h-41.8V79.1z M124.3,172.4h24.5c34.9,0,42.9-26.5,42.9-39.7c0-21.5-13.7-39.7-43.7-39.7h-23.7V172.4z}
                 svg{M88.7,56.8c0,5.5-4.5,10.1-10.1,10.1c-5.6,0-10.1-4.6-10.1-10.1c0-5.6,4.5-10.1,10.1-10.1C84.2,46.7,88.7,51.3,88.7,56.8z};
  }
}
\newcommand\orcidicon[1]{\href{https://orcid.org/#1}{\mbox{\scalerel*{
\begin{tikzpicture}[yscale=-1,transform shape]
\pic{orcidlogo};
\end{tikzpicture}
}{|}}}}